\documentclass{article}
\usepackage{ijcai25}

\usepackage{times}
\usepackage{soul}
\usepackage{url}
\usepackage[hidelinks]{hyperref}
\usepackage[utf8]{inputenc}
\usepackage[small]{caption}
\usepackage{graphicx}
\usepackage{amsmath}
\usepackage{amsthm}
\usepackage{booktabs}
\usepackage{algorithm}
\usepackage{algorithmic}
\usepackage[switch]{lineno}

\usepackage{adjustbox}
\usepackage{multirow}
\usepackage{colortbl}
\usepackage{tcolorbox} 
\usepackage{xcolor}   

\usepackage{markdown}   
\usepackage{listings}  
\usepackage{enumitem}   
\usepackage{fancyvrb}
\usepackage{multicol} 
\tcbuselibrary{breakable}  
\usepackage{enumitem}  
\usepackage{hyperref}  
\usepackage{fvextra}   
\DefineVerbatimEnvironment{Verbatim}{Verbatim}{breaklines=true, breakanywhere=true}

\title{MIRROR: Multi-agent Intra- and Inter-Reflection for Optimized Reasoning in Tool Learning}

\author{
Zikang Guo$^1$ \thanks{Work done during the internship at Metastone.}
\and
Benfeng Xu$^{1}$\and
Xiaorui Wang$^2$\And
Zhendong Mao$^1$\thanks{Corresponding author: Zhendong Mao.}\\
\affiliations
$^1$University of Science and Technology of China\\
$^2$Metastone Technology, Beijing, China\\
\emails
\texttt{\{gzk170401, benfeng\}@mail.ustc.edu.cn\\
harrywxr@outlook.com,
zdmao@ustc.edu.cn
}}

\begin{document}

\maketitle

\begin{abstract}
Complex tasks involving tool integration pose significant challenges for Large Language Models (LLMs), leading to the emergence of multi-agent workflows as a promising solution.
Reflection has emerged as an effective strategy for correcting erroneous trajectories in agentic workflows. However, existing approaches only exploit such capability in the post-action stage, where the agent observes the execution outcomes. We argue that, like humans, LLMs can also engage in reflection before action execution: the agent can \textit{anticipate} undesirable outcomes from its own decisions, which not only provides a necessarily complementary perspective to evaluate the decision but also prevents the propagation of errors throughout the trajectory.
In this paper, we propose \textbf{MIRROR}, a framework that consists of both \textbf{intra-reflection}, which critically assesses intended actions before execution, and \textbf{inter-reflection}, which further adjusts the trajectory based on observations. This design systematically leverages LLM reflection capabilities to eliminate and rectify erroneous actions on a more comprehensive scope.
Evaluations on both the StableToolBench and TravelPlanner benchmarks demonstrate MIRROR's superior performance, achieving state-of-the-art results compared to existing approaches.

\end{abstract}

\section{Introduction}

\begin{figure}[t]
\centering
\includegraphics[width=0.99\linewidth]{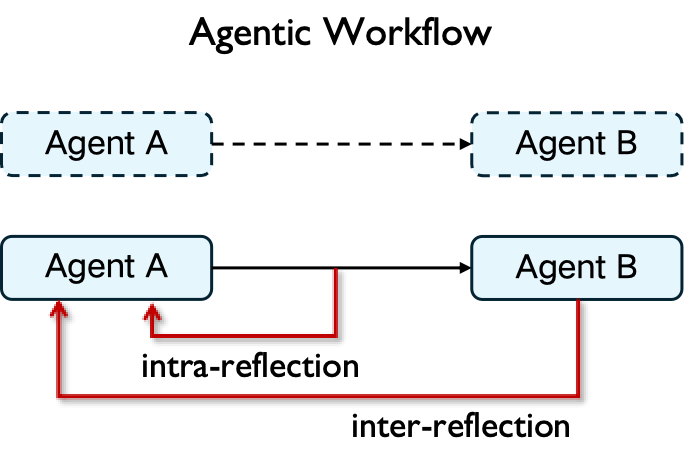}
\caption{Architectural comparison of intra-reflection and inter-reflection in MIRROR. Intra-reflection operates as a preventive mechanism within each agent, evaluating decisions before execution to prevent errors. Inter-reflection functions as a corrective mechanism between agents, enabling system-wide optimization through post-execution learning. This dual-phase approach creates a comprehensive error prevention and correction system.}
\label{fig:workflow}
\end{figure}

Large Language Models (LLMs) have demonstrated remarkable capabilities in natural language processing, showcasing unprecedented performance in planning, reasoning, and complex task completion~\cite{brown2020:language,chen2022:program,sun2023survey,yang2023foundation,touvron2023llama}. However, these models face fundamental limitations in accessing real-time information, processing novel data patterns, and executing precise system controls \cite{petroni2020kilt,lewis2020retrieval,komeili2021internet,zhang2022opt}. These constraints become particularly challenging when tackling real-world tasks that require dynamic adaptation and accurate tool manipulation.

The tool learning paradigm emerged as a natural solution to extend LLMs' capabilities beyond their inherent limitations \cite{qin2023toolllm}. By enabling LLMs to interact with external tools and APIs, this approach significantly expands their practical applications. 
Recent advances in foundation models have further enhanced their ability to handle basic tool interactions \cite{achiam2023gpt,team2023gemini,dubey2024llama,qwen2.5}. However, when confronted with sophisticated multi-step tasks requiring intricate tool coordination, even these advanced models struggle to maintain consistent performance.

This challenge has given rise to multi-agent workflows, where complex tasks are distributed across specialized agents \cite{wu2023autogen,hong2024metagpt,opendevin}. In these systems, reflection mechanisms play a crucial role in optimizing agent behavior. Current approaches, exemplified by Reflexion \cite{shinn2024reflexion}, implement reflection as a post-execution analysis tool, where agents learn from completed actions. While this retrospective approach enables iterative improvement, it suffers from fundamental limitations: it cannot prevent initial errors, risks irreversible system changes, and incurs significant learning costs through trial and error.

To address these limitations, we propose intra-reflection: a proactive evaluation mechanism that occurs before action execution. This approach mirrors human cognitive processes, where we often mentally simulate outcomes before taking action. By combining this preventive strategy with traditional post-execution reflection, we create a more comprehensive framework for multi-agent learning and adaptation.

Our paper introduces \textbf{MIRROR} (Multi-agent Intra- and Inter-Reflection for Optimized Reasoning), a novel framework that seamlessly integrates both reflection phases. This dual-phase approach systematically prevents errors while optimizing learning from necessary failures, creating a more robust and efficient multi-agent system. Unlike existing solutions that rely solely on learning from mistakes, MIRROR enables agents to anticipate and avoid potential errors while maintaining the ability to learn from unavoidable failures.

The primary contributions of this paper are summarized as follows:

\begin{itemize}
\item We introduce the concept of intra-reflection in multi-agent tool learning, a pre-action evaluation mechanism that allows agents to critically assess their intended outputs before execution. 
\item We propose MIRROR, a novel framework for multi-agent systems in tool learning tasks. MIRROR employs both intra-reflection and inter-reflection mechanisms to enhance decision-making, prevent error propagation, and improve collaboration efficiency. 
\item We provide a comprehensive evaluation of our framework using both StableToolBench and TravelPlanner benchmarks, demonstrating its superior performance across evaluation metrics, significantly outperforming existing state-of-the-art approaches.
\end{itemize}

\section{Related Work}
\paragraph{LLM-Based Task Planning.} 
LLMs have been increasingly leveraged for task planning. Initial approaches like Chain-of-Thought (CoT) \cite{wei2022chain} generate intermediate reasoning steps, while Self-Consistency \cite{wang2022self} improves robustness by generating multiple reasoning paths and selecting the most consistent solution. ReAct \cite{yao2022react} integrates reasoning with action-taking capabilities. More advanced methods explore complex reasoning structures: Tree-of-Thought (ToT) \cite{yao2024tree} enables simultaneous exploration and evaluation of multiple reasoning paths, and Graph of Thoughts (GoT) \cite{besta2024graph} provides even greater flexibility with graph-structured, multi-directional thought connections. Our method builds on these by incorporating intra-reflection, allowing LLMs to critically assess and refine reasoning during task execution for enhanced adaptability.

\paragraph{Reflection in LLMs.} 
LLM frameworks increasingly integrate reflection mechanisms to enhance performance and adaptability. For example, Self-refine \cite{madaan2024self} employs iterative self-critique and refinement, while Reflexion \cite{shinn2024reflexion} incorporates self-monitoring and feedback loops within the ReAct framework. DFSDT \cite{qin2023toolllm} advances reflection by enabling models to re-evaluate and re-plan based on their complete history of previous error paths and information, thereby broadening the explored solution space. CRITIC \cite{gou2023critic} leverages external tools for output validation and subsequent improvement. While these approaches predominantly emphasize inter-reflection or post-hoc refinement, MIRROR innovates by stressing continuous intra-reflection during all reasoning phases, facilitating more dynamic and context-aware adaptations in real time.

\paragraph{LLM-Powered Multi-Agent Systems.} 
Single-agent LLM systems face limitations such as hallucinations, restricted context, and lack of collaboration in real-world applications \cite{huang2024understanding}. LLM-powered multi-agent systems (MAS) offer a promising solution, where multiple autonomous agents with unique strategies collaborate to manage complex, dynamic tasks beyond single-agent capabilities \cite{zhuge2023mindstorms}. This collaborative approach leverages distinct agent strengths through communication. Notable examples include AutoGen \cite{wu2023autogen}, MetaGPT \cite{hong2024metagpt}, and OpenHands \cite{opendevin}, which utilize Agent-Computer Interfaces for enhanced problem-solving. Our approach extends these MAS frameworks by integrating both intra-reflection and inter-reflection mechanisms, aiming to better address real-world complexities through maximized collaborative potential and adaptive reasoning.

\section{Methodology}

\begin{figure*}[t]
\centering
\includegraphics[width=0.98\linewidth]{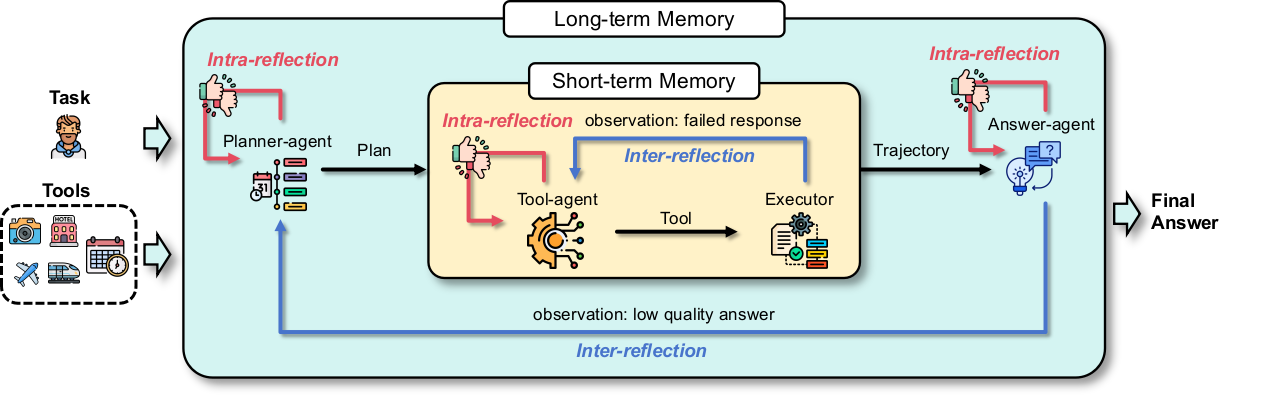}
\caption{Overview of the MIRROR framework. Given a task and candidate tools, the Planner Agent decomposes the complex task into subtasks. The Tool Agent selects appropriate tools and parameters for each, which are passed to the Executor for execution. The Answer Agent then utilizes the entire execution trajectory to produce the final answer. Critically, each agent undergoes intra-reflection before passing its output, and a dual-memory architecture facilitates inter-reflection.}
\label{fig: framework}
\end{figure*}

The MIRROR framework presents a structured approach to enhance LLM capabilities in tool learning. It leverages a specialized multi-agent system synergistically combined with unique intra-reflection and inter-reflection mechanisms. The overarching goal is to optimize the decomposition of complex tasks, the selection and parameterization of tools, and the final generation of answers, while proactively minimizing errors and enabling continuous learning.

\subsection{Task Definition}
Tool learning tasks involve an LLM interacting with external tools/APIs. Given a task description ($Q$) and a tool library ($T = \{n, d, p\}$, where $n$ is name, $d$ is description, $p$ is parameters), the system must autonomously select the optimal tool ($t \in T$) and its parameters ($p_t$). These are passed to an Executor, and the resulting output is synthesized into a comprehensive answer.

\subsection{Framework Overview}
MIRROR's core innovation is \textbf{intra-reflection}, a distinct contribution enabling meaningful self-evaluation \textit{within a single agent during its generation phase and before action execution or handoff to another agent}. This contrasts with methods relying on external agent feedback or post-execution reflection. Intra-reflection functions as a formalized prompting technique, universally applicable across tasks and models, embedding evaluation criteria directly into the agent's prompt to guide its reasoning and output generation. This systematic approach facilitates proactive error prevention. This overall framework, which deeply embeds intra-reflection within each agent and also integrates inter-reflection for comprehensive learning, is visually outlined in Figure~\ref{fig: framework}.

\subsection{Multi-Agent System}
MIRROR integrates three distinct, specialized agents:

\paragraph{Planner Agent:} Decomposes complex tasks into a topologically structured sequence of manageable subtasks, considering dependencies and execution order. It utilizes Long-Term Memory to refine decomposition strategies based on outcomes from previous attempts within the current task.

\paragraph{Tool Agent:} Selects appropriate tools and parameters for each subtask. Its decision-making is guided by Short-Term Memory, which includes execution failure information and the agent's own intra-reflection outputs from prior attempts on the current subtask, helping to avoid repeated errors. Successful subtask results are stored for subsequent use.

\paragraph{Answer Agent:} Synthesizes individual results from all completed subtasks, considering the entire execution trajectory, to produce a coherent and comprehensive final answer to the original task.

These agents engage in collaborative interactions, where the validated output of one agent becomes the input for the next. This structured collaboration, augmented by reflection, aims to achieve solutions that surpass the capabilities of any single agent operating in isolation.

\subsection{Intra-Reflection}
A cornerstone of the MIRROR framework, the intra-reflection process is implemented by each agent \textit{before} its output is passed to the next agent or the Executor. This ensures a thorough self-evaluation, minimizing errors and optimizing the quality of inter-agent information flow. Each agent performs a self-assessment based on its specific role and predefined quality thresholds ($\theta_p, \theta_t, \theta_a$). This mechanism does not depend on additional LLMs or external evaluation agents, functioning as an inherent part of the agent's generation cycle.

\paragraph{Planner Agent Intra-Reflection:} After generating an initial task decomposition, the Planner Agent critically evaluates this plan. The assessment focuses on its completeness, efficiency, and coherence. It also considers if subtasks can be further optimized. The agent integrates lessons from prior experiences within the current task to enhance decomposition quality. If the internal evaluation score surpasses the threshold $\theta_p$, the plan is deemed sufficiently robust and proceeds to the Tool Agent. Otherwise, the plan is iteratively revised until it meets the required quality standard. This iterative refinement ensures that only well-optimized and coherent plans advance, significantly improving the framework's overall effectiveness and resource utilization.

\paragraph{Tool Agent Intra-Reflection:} Upon receiving a subtask and selecting a tool and parameters, the Tool Agent performs intra-reflection. It evaluates these choices based on how well they meet the specific subtask requirements, whether they incorporate lessons from past attempts (within the current task, leveraging STM), and their alignment with the subtask's objectives. The agent assigns a score reflecting the expected effectiveness of its choices. Selections scoring above threshold $\theta_t$ are passed to the Executor for execution. Those scoring below are revised (the agent may choose a different tool or adjust parameters) to better align with subtask goals. This process optimizes tool and strategy selection at a granular level, enhancing the framework's operational efficiency and increasing the probability of successful subtask completion.

\paragraph{Answer Agent Intra-Reflection:} After synthesizing the results from all subtasks into a comprehensive solution, the Answer Agent conducts a final intra-reflection. This evaluation focuses on the completeness of the answer, its integration of key actions and observations from the entire trajectory, and the clarity of its presentation. The agent assigns a score to its output. If this score meets or exceeds the threshold $\theta_a$, the task is considered successfully completed. If the score falls below $\theta_a$, indicating a suboptimal answer, the entire task trajectory (including the plan, tool uses, reflections, and the deficient answer) is stored in long-term memory. This stored information is then utilized by the Planner Agent to refine its strategy for the current task, facilitating an improved attempt at successful completion and promoting continuous, task-specific adaptation.

\subsection{Inter-Reflection}
\label{dual-memory}

MIRROR employs a dual-memory architecture to facilitate inter-reflection:

\paragraph{Short-Term Memory (STM):} Utilized by the Tool Agent for subtask tool/parameter selection. It records execution failures and the agent's intra-reflection outputs, prompting adjustments. STM is reset upon successful subtask completion to maintain relevance for new selections.

\paragraph{Long-Term Memory (LTM):} This memory is \textbf{task-specific}. It stores the entire trajectory for the current task (decomposition plan, tool selections, execution results, solution quality). LTM is updated if execution fails or if the Answer Agent's final output is below $\theta_a$. It resets upon overall task completion, enabling the system to learn from successes and failures \textit{within the current task} for progressive strategy refinement.

This dual-memory architecture allows MIRROR to continually adapt and refine its performance through structured experiential learning.

\subsection{Agent Collaboration}
Agent collaboration critically relies on intra-reflection for high-quality inter-agent handoffs. The Planner Agent passes decompositions, validated for coherence and feasibility by its intra-reflection, to the Tool Agent. The Tool Agent, after its own reflection-refined tool and parameter selection aimed at optimal subtask execution, then directs the process. Results subsequently flow to the Answer Agent for careful synthesis and final quality assurance, also guided by its intra-reflection.

This reflection-gated workflow minimizes error propagation via early self-correction and optimizes inter-agent communication with reliable information exchange. Inter-reflection, through dual memory, further enhances collaboration: STM aids Tool Agent's immediate adaptation to execution challenges, while LTM enables the Planner Agent to refine strategies from current task trajectories.

Integrating both reflection types creates a self-optimizing system. This enhances individual agent efficiency and overall framework performance, robustness, and adaptability for complex tasks. Thus, MIRROR continuously learns and evolves. While dual-reflection increases token processing, this is justified by significant, experimentally validated performance gains, making it a worthwhile trade-off.

\section{Experiment}
To evaluate our proposed framework, we conducted comprehensive experiments as detailed below.

\subsection{Setup}
\paragraph{Benchmarks}
We utilized two benchmarks:
\begin{itemize}
\item \textbf{StableToolBench} \cite{guo2024stabletoolbench}: An extension of ToolBench \cite{qin2023toolllm}, this benchmark assesses LLM function invocation capabilities through executable tests.
\item \textbf{TravelPlanner} \cite{xie2024travelplanner}: This benchmark evaluates agent proficiency in real-world planning and tool interactions. Due to computational constraints, we used its validation set in a two-stage mode to assess tool utilization and planning capabilities.
\end{itemize}

\paragraph{Metrics}
\begin{itemize}
\item \textbf{StableToolBench}: Metrics include Pass Rate (percentage of tasks successfully completed) and Win Rate (comparison against GPT-3.5+ReAct via LLM evaluation).
\item \textbf{TravelPlanner}: Metrics include Delivery Rate (plan completion), Commonsense and Hard Constraint Pass Rates (rule adherence), and Final Pass Rate (overall plan feasibility).
\end{itemize}

\paragraph{Model} Our experiments used the following foundational LLMs: GPT-3.5 Turbo \cite{ouyang2022training} (\textsl{gpt-35-turbo-0125}), GPT-4o Mini \cite{achiam2023gpt} (\textsl{gpt-4o-mini-2024-07-18}), GPT-4o \cite{achiam2023gpt} (\textsl{gpt-4o-2024-05-13}), Claude 3 Haiku \cite{anthropic_claude3}, and Qwen2.5-72B \cite{qwen2.5}. For evaluation consistency on StableToolBench, we employed GPT-4 Turbo \cite{achiam2023gpt} (\textsl{gpt-4-turbo-2024-04-09}). GPT-4o served as the parsing model for TravelPlanner.

\paragraph{Baselines}
We compared our method against six baselines:
\begin{itemize}
\item \textbf{ReAct} \cite{yao2022react}: Alternates reasoning and acting within an LLM for interpretable trace generation and task-specific action execution.
\item \textbf{DFSDT} \cite{qin2023toolllm}: Constructs and traverses decision trees for systematic re-evaluation and re-planning based on previous error paths.
\item \textbf{Reflexion} \cite{shinn2024reflexion}: Employs self-reflection on environmental interactions to improve decision-making.
\item \textbf{Smurfs} \cite{chen2024smurfs}: A multi-agent framework assigning distinct roles to agents via prompting for collaborative problem-solving.
\item \textbf{ToolLlama-2} \cite{qin2023toolllm}: A model fine-tuned on the ToolBench dataset.
\item \textbf{ToolGen} \cite{wang2024toolgen}: Represents tools as unique vocabulary tokens for direct tool call generation via a three-stage training process, bypassing separate retrieval steps.
\end{itemize}

\begin{table*}[t]
\centering
\footnotesize
\begin{adjustbox}{width=0.98\textwidth}
\setlength{\tabcolsep}{4pt}
\begin{tabular}{@{}ll|cc|cc|cc|cc|cc|cc|cc@{}}
\toprule
\textbf{LLM Core} & \textbf{Method} & \multicolumn{2}{c|}{\textbf{G1-Inst.}} & \multicolumn{2}{c|}{\textbf{G1-Cat.}} & \multicolumn{2}{c|}{\textbf{G1-Tool.}} & \multicolumn{2}{c|}{\textbf{G2-Inst.}} & \multicolumn{2}{c|}{\textbf{G2-Cat.}} & \multicolumn{2}{c|}{\textbf{G3-Inst.}} & \multicolumn{2}{c}{\textbf{Avg.}} \\
 &  & Pass. & Win. & Pass. & Win. & Pass. & Win. & Pass. & Win. & Pass. & Win. & Pass. & Win. & Pass. & Win. \\
\midrule
\multirow{5}{*}{GPT-3.5 Turbo} & ReAct & 47.2\textsubscript{±0.5}& / & 42.9\textsubscript{±0.8}& / & 41.8\textsubscript{±0.5}& / & 46.1\textsubscript{±0.2}& / & 43.1\textsubscript{±0.3}& / & 42.1\textsubscript{±1.5}& / & 43.9\textsubscript{±0.5}& / \\
 & DFSDT & 64.5\textsubscript{±0.4}& 66.9 & 61.2\textsubscript{±0.8}& 64.7 & 64.7\textsubscript{±0.7}& 68.4 & 59.3\textsubscript{±0.2}& 60.4 & 61.6\textsubscript{±1.2}& 64.5 & 57.4\textsubscript{±1.3}& 59.0 & 61.5\textsubscript{±0.8}& 64.0 \\
 & Reflexion & 75.8\textsubscript{±1.0}& 65.6 & 73.0\textsubscript{±1.5}& 76.5 & 75.3\textsubscript{±1.2}& 70.3 & 72.5\textsubscript{±2.8}& 72.6 & 73.0\textsubscript{±0.6}& 72.6 & 58.5\textsubscript{±0.8}& 49.2 & 71.4\textsubscript{±1.3}& 67.8 \\
 & Smurfs & 75.9\textsubscript{±1.1}& 74.2 & 82.6\textsubscript{±0.4}& 81.7 & 72.2\textsubscript{±0.9}& 71.5 & 76.6\textsubscript{±1.4}& 73.6 & 79.6\textsubscript{±0.2}& 79.8 & 73.0\textsubscript{±2.0}& 68.9 & 76.7\textsubscript{±1.0}& 75.0 \\
 &\cellcolor{green!18}\textbf{MIRROR}&\cellcolor{green!18}\textbf{82.5\textsubscript{±0.5}}& \cellcolor{green!18}\textbf{76.1} &\cellcolor{green!18}\textbf{86.3\textsubscript{±1.3}}& \cellcolor{green!18}\textbf{87.6} & \cellcolor{green!18}\textbf{84.9\textsubscript{±0.6}}& \cellcolor{green!18}\textbf{79.7} & \cellcolor{green!18}\textbf{78.8\textsubscript{±1.0}}& \cellcolor{green!18}\textbf{87.7} & \cellcolor{green!18}\textbf{82.4\textsubscript{±1.0}}& \cellcolor{green!18}\textbf{83.1} & \cellcolor{green!18}\textbf{87.4\textsubscript{±1.5}}& \cellcolor{green!18}\textbf{80.3} & \cellcolor{green!18}\textbf{83.7\textsubscript{±1.0}}& \cellcolor{green!18}\textbf{82.4} \\
\midrule
\multirow{5}{*}{Claude 3 Haiku} & ReAct & 54.9\textsubscript{±0.0} & 49.1 & 46.1\textsubscript{±0.0} & 47.7 & 54.0\textsubscript{±0.6} & 55.1 & 55.3\textsubscript{±1.1} & 52.8 & 52.2\textsubscript{±1.0} & 50.8 & 65.8\textsubscript{±0.8} & 55.7 & 54.7\textsubscript{±0.6} & 51.9 \\
& DFSDT & 67.9\textsubscript{±0.8} & 55.2 & 68.8\textsubscript{±0.9} & 65.4 & 76.8\textsubscript{±0.6} & 68.4 & 65.9\textsubscript{±1.6} & 67.9 & 69.8\textsubscript{±0.9} & 57.3 & 69.4\textsubscript{±0.8} & 57.4 & 69.8\textsubscript{±0.9} & 61.9 \\
& Reflexion & 75.8\textsubscript{±0.7} & 68.7 & 76.9\textsubscript{±0.4} & 85.0 & 80.6\textsubscript{±0.4} & 77.2 & 73.4\textsubscript{±1.1} & 77.4 & 73.8\textsubscript{±1.3} & 76.6 & 73.0\textsubscript{±1.2} & 70.5 & 75.6\textsubscript{±0.9} & 75.9 \\
& Smurfs & 68.4\textsubscript{±0.5} & 65.0 & 81.2\textsubscript{±1.2} & 81.0 & 80.6\textsubscript{±1.3} & 76.6 & 72.5\textsubscript{±1.2} & 71.7 & 75.4\textsubscript{±0.3} & 69.4 & 74.3\textsubscript{±0.8} & 60.7 & 75.4\textsubscript{±0.9} & 70.7 \\
 &\cellcolor{green!18}\textbf{MIRROR} & \cellcolor{green!18}\textbf{78.3\textsubscript{±0.8}} & \cellcolor{green!18}\textbf{74.8} & \cellcolor{green!18}\textbf{81.9\textsubscript{±0.6}} & \cellcolor{green!18}\textbf{86.9} & \cellcolor{green!18}\textbf{85.2\textsubscript{±0.8}} & \cellcolor{green!18}\textbf{79.7} & \cellcolor{green!18}\textbf{80.2\textsubscript{±1.7}} & \cellcolor{green!18}\textbf{81.1} & \cellcolor{green!18}\textbf{78.1\textsubscript{±1.4}} & \cellcolor{green!18}\textbf{81.5} & \cellcolor{green!18}\textbf{81.4\textsubscript{±0.4}} & \cellcolor{green!18}\textbf{75.4} & \cellcolor{green!18}\textbf{80.9\textsubscript{±1.0}} & \cellcolor{green!18}\textbf{79.9} \\
\midrule
\multirow{5}{*}{Qwen2.5-72B} & ReAct & 53.0\textsubscript{±1.0} & 52.8 & 63.6\textsubscript{±0.4} & 69.3 & 60.8\textsubscript{±0.8} & 62.0 & 60.8\textsubscript{±0.7} & 67.9 & 57.5\textsubscript{±0.7} & 60.5 & 54.9\textsubscript{±2.7} & 52.5 & 58.4\textsubscript{±1.1} & 60.8 \\
& DFSDT & 62.9\textsubscript{±0.4} & 60.7 & 66.3\textsubscript{±0.7} & 72.5 & 62.4\textsubscript{±0.1} & 62.7 & 64.0\textsubscript{±1.2} & 66.0 & 65.3\textsubscript{±0.9} & 64.5 & 58.5\textsubscript{±1.4} & 52.5 & 63.2\textsubscript{±0.8} & 63.2 \\
& Reflexion & 76.4\textsubscript{±0.3} & 70.6 & 75.7\textsubscript{±0.6} & 81.0 & 75.9\textsubscript{±0.9} & 77.2 & 72.3\textsubscript{±0.2} & 83.0 & 77.2\textsubscript{±0.7} & 81.5 & 79.0\textsubscript{±0.8} & 68.9 & 76.1\textsubscript{±0.6} & 77.0 \\
& Smurfs & 77.8\textsubscript{±0.9} & 73.6 & 81.9\textsubscript{±0.2} & 88.9 & 78.0\textsubscript{±0.1} & 78.5 & 76.7\textsubscript{±0.6} & 84.0 & 79.6\textsubscript{±0.7} & 77.4 & 88.3\textsubscript{±0.4} & 68.9 & 80.4\textsubscript{±0.5} & 78.6 \\
&\cellcolor{green!18}\textbf{MIRROR} & \cellcolor{green!18}\textbf{83.2\textsubscript{±0.8}} & \cellcolor{green!18}\textbf{75.5} & \cellcolor{green!18}\textbf{85.3\textsubscript{±0.3}} & \cellcolor{green!18}\textbf{90.8} & \cellcolor{green!18}\textbf{81.8\textsubscript{±0.3}} & \cellcolor{green!18}\textbf{80.4} & \cellcolor{green!18}\textbf{77.2\textsubscript{±0.3}} & \cellcolor{green!18}\textbf{86.8} & \cellcolor{green!18}\textbf{80.9\textsubscript{±1.1}} & \cellcolor{green!18}\textbf{83.9} & \cellcolor{green!18}\textbf{89.1\textsubscript{±0.8}} & \cellcolor{green!18}\textbf{85.2} & \cellcolor{green!18}\textbf{82.9\textsubscript{±0.6}} & \cellcolor{green!18}\textbf{83.8} \\
\midrule
\multirow{5}{*}{GPT-4o} & ReAct & 58.6\textsubscript{±1.1} & 63.8 & 61.5\textsubscript{±0.7} & 64.1 & 62.2\textsubscript{±0.3} & 59.5 & 61.5\textsubscript{±1.6} & 67.0 & 54.7\textsubscript{±1.9} & 62.9 & 52.5\textsubscript{±0.7} & 50.8 & 58.5\textsubscript{±1.1} & 61.4 \\
 & DFSDT & 68.7\textsubscript{±1.9} & 66.3 & 66.1\textsubscript{±0.7} & 67.3 & 73.9\textsubscript{±0.5} & 72.8 & 67.0\textsubscript{±1.0} & 69.8 & 65.9\textsubscript{±0.7} & 74.2 & 67.5\textsubscript{±1.0} & 72.1 & 68.2\textsubscript{±1.0} & 70.4 \\
  & Reflexion & 76.4\textsubscript{±0.3} & 74.8 & 76.7\textsubscript{±0.9} & 87.6 & 74.4\textsubscript{±0.4} & 78.5 & 71.5\textsubscript{±0.8} & 78.3 & 73.5\textsubscript{±1.1} & 79.0 & 72.7\textsubscript{±0.8} & 70.5 & 74.2\textsubscript{±0.7} & 78.1 \\
 & Smurfs & 75.1\textsubscript{±1.3} & 73.6 & 79.5\textsubscript{±1.1} & 80.4 & 81.6\textsubscript{±1.2} & 76.6 & 77.0\textsubscript{±0.4} & 83.0 & 76.9\textsubscript{±1.0} & 78.2 & 84.4\textsubscript{±0.7} & 78.7 & 79.1\textsubscript{±1.0} & 78.4 \\
 & \cellcolor{green!18}\textbf{MIRROR} & \cellcolor{green!18}\textbf{82.5\textsubscript{±0.3}} & \cellcolor{green!18}\textbf{77.3} & \cellcolor{green!18}\textbf{83.3\textsubscript{±0.5}} & \cellcolor{green!18}\textbf{91.5} & \cellcolor{green!18}\textbf{86.2\textsubscript{±0.7}} & \cellcolor{green!18}\textbf{86.1} & \cellcolor{green!18}\textbf{77.2\textsubscript{±0.2}} & \cellcolor{green!18}\textbf{84.9} & \cellcolor{green!18}\textbf{80.8\textsubscript{±1.0}} & \cellcolor{green!18}\textbf{85.5} & \cellcolor{green!18}\textbf{88.3\textsubscript{±1.5}} & \cellcolor{green!18}\textbf{86.9} & \cellcolor{green!18}\textbf{83.1\textsubscript{±0.7}} & \cellcolor{green!18}\textbf{85.4} \\
 \midrule
 \multirow{1}{*}{ToolLlama-2} & SFT & 54.0\textsubscript{±0.9} & 48.5 & 54.1\textsubscript{±0.4} & 50.3 & 46.1\textsubscript{±2.1} & 49.4 & 37.9\textsubscript{±1.5} & 46.2 & 43.1\textsubscript{±0.7} & 46.8 & 41.3\textsubscript{±0.8} & 37.7 & 46.1\textsubscript{±1.1} & 46.5 \\
 \midrule
  \multirow{1}{*}{ToolGen} & SFT & 51.2\textsubscript{±0.9} & 47.9 & 47.8\textsubscript{±0.6} & 45.1 & 52.4\textsubscript{±0.3} & 43.7 & 48.7\textsubscript{±2.0} & 52.8 & 42.3\textsubscript{±0.3} & 45.2 & 35.5\textsubscript{±1.0} & 29.5 & 46.3\textsubscript{±0.9} & 44.0 \\
\bottomrule
\end{tabular}
\end{adjustbox}
\caption{Main results on the StableToolBench benchmark. The best-performing method under each LLM core is highlighted in \textbf{bold}. Pass Rate (\%) and Win Rate (\%) are the evaluation metrics, with Win Rate calculated by comparing each method to GPT-3.5+ReAct.}
\label{tab:stable-tool-bench}
\end{table*}

\begin{table}[t]
\centering
\small
\footnotesize
\setlength{\tabcolsep}{2pt}
\resizebox{\columnwidth}{!}{
\scriptsize
\setlength{\tabcolsep}{2pt}
\begin{tabular}{@{}ll|c|cc|cc|c@{}}
\toprule
\textbf{LLM Core} & \textbf{Method} & \textbf{Delivery} & \multicolumn{2}{c|}{\textbf{Commonsense}} & \multicolumn{2}{c|}{\textbf{Hard Constraint}} & \textbf{Final} \\
& & \textbf{Rate} & \textbf{Micro} & \textbf{Macro} & \textbf{Micro} & \textbf{Macro} & \textbf{Pass Rate} \\
\midrule
\multirow{2}{*}{GPT-4o Mini} & ReAct & 75.0 & 54.7 & 0 & 0 & 0 & 0 \\
& \textbf{MIRROR} &\textbf{96.1} & \textbf{69.0} & \textbf{3.3} & \textbf{5.2} & \textbf{2.2} & 0 \\
\midrule
\multirow{2}{*}{Qwen2.5-72B} & ReAct & 79.4 & 56.0 & 0.6 & 3.3 & 2.2 & 0 \\
& \textbf{MIRROR} & \textbf{95.6} & \textbf{64.9} & \textbf{6.7} & \textbf{12.9} & \textbf{9.4} & \textbf{1.7} \\
\midrule
\multirow{2}{*}{GPT-4o} & ReAct & 85.6 & 61.4 & 5.0 & 15.2 & 7.8 & 2.2 \\
& \textbf{MIRROR} & \textbf{100} & \textbf{73.2} & \textbf{13.9} & \textbf{27.6} & \textbf{13.3} & 2.2 \\
\bottomrule
\end{tabular}
}
\caption{Evaluation results on TravelPlanner benchmark. MIRROR consistently outperforms ReAct across different LLM cores.}
\label{tab:travel-planner}
\end{table}

\subsection{Main Results}
\paragraph{Experimental results demonstrate MIRROR's consistent performance advantages across both benchmarks.} 
Across both benchmarks, MIRROR surpasses existing baselines. On StableToolBench, it demonstrates superiority with diverse LLMs (open-source and proprietary), notably outperforming supervised fine-tuning (SFT) methods like ToolLlama-2 and ToolGen (both with Pass Rates around 46\%), underscoring SFT's limitations in complex tool use. This superior performance extends to TravelPlanner, where MIRROR consistently leads ReAct in Delivery Rate, Commonsense Pass Rate, and Hard Constraint Pass Rate. The efficacy of MIRROR is attributed to its dual-reflection architecture, which incorporates intra-reflection for immediate error detection and inter-reflection for strategic optimization. This approach shows clear advantages over Reflexion's single inter-reflection mechanism. Despite these gains, achieving a high Final Pass Rate on TravelPlanner remains challenging due to the inherent complexity of managing information and constraints in multifaceted planning.

\paragraph{Performance across different LLM cores.} 
MIRROR's effectiveness varies with the LLM core and benchmark.
Across both StableToolBench (Table \ref{tab:stable-tool-bench}) and TravelPlanner (Table \ref{tab:travel-planner}), MIRROR consistently enhanced performance over baselines across all evaluated LLMs.
On StableToolBench, this led to average Pass Rate gains (over the next best method) from \textbf{2.5\%} (Qwen2.5-72B) to \textbf{7.0\%} (GPT-3.5 Turbo), and Win Rate gains (vs. GPT-3.5+ReAct) from \textbf{5.2\%} (Qwen2.5-72B) to \textbf{9.2\%} (Claude 3 Haiku).
On TravelPlanner, which tests complex multi-step planning, MIRROR increased Delivery Rates by amounts ranging from \textbf{14.4\%} (with GPT-4o) to \textbf{21.1\%} (with GPT-4o Mini). Further, it yielded absolute improvements in macro Pass Rates for Commonsense (from \textbf{3.3\%} with GPT-4o Mini to \textbf{8.9\%} with GPT-4o) and Hard Constraints (from \textbf{2.2\%} with GPT-4o Mini to \textbf{7.2\%} with Qwen2.5-72B).

\subsection{Ablation Study}

\begin{table*}[ht]
\centering
\adjustbox{max width=\linewidth}{
\begin{tabular}{@{}lccccccc@{}} 
\toprule
\textbf{Configuration} & \textbf{G1-Inst.} & \textbf{G1-Cat.} & \textbf{G1-Tool.} & \textbf{G2-Inst.} & \textbf{G2-Cat.} & \textbf{G3-Inst.} & \textbf{Avg.} \\
\midrule
\textbf{MIRROR} & \textbf{84.9\textsubscript{±1.2}} & \textbf{87.9\textsubscript{±0.5}} & \textbf{89.5\textsubscript{±0.8}} & \textbf{82.7\textsubscript{±0.6}} & \textbf{83.1\textsubscript{±0.7}} & \textbf{86.1\textsubscript{±0.7}} & \textbf{85.7\textsubscript{±0.8}} \\
\hspace{5mm} FC & 78.6\textsubscript{±0.4} & 83.7\textsubscript{±0.5} & 79.3\textsubscript{±0.3} & 80.2\textsubscript{±0.7} & 85.6\textsubscript{±2.2} & 91.8\textsubscript{±0.7} & 83.2\textsubscript{±0.8}  \\
\midrule 
\hspace{5mm} w/o Planner Intra. & 83.2\textsubscript{±0.5} & 83.3\textsubscript{±0.3} & 87.0\textsubscript{±0.7} & 76.7\textsubscript{±1.1} & 81.5\textsubscript{±0.9} & 80.9\textsubscript{±1.4} & 82.1\textsubscript{±0.8} \\
\hspace{5mm} w/o Tool Intra. & 80.4\textsubscript{±0.0} & 84.6\textsubscript{±0.5} & 83.3\textsubscript{±0.8} & 79.7\textsubscript{±0.4} & 80.8\textsubscript{±0.5} & 79.0\textsubscript{±0.4} & 81.3\textsubscript{±0.4} \\
\hspace{5mm} w/o Answer Intra. & 81.7\textsubscript{±0.6} & 80.8\textsubscript{±0.6} & 83.1\textsubscript{±0.9} & 74.8\textsubscript{±0.8} & 78.6\textsubscript{±1.0} & 77.6\textsubscript{±1.7} & 79.4\textsubscript{±0.9} \\
\hspace{5mm} w/o Intra. & 78.0\textsubscript{±1.1} & 79.0\textsubscript{±0.3} & 81.3\textsubscript{±1.2} & 76.7\textsubscript{±0.4} & 80.8\textsubscript{±0.8} & 76.2\textsubscript{±1.3} & 78.7\textsubscript{±0.9} \\
\midrule
\hspace{5mm} w/o Long-term Memory & 83.5\textsubscript{±0.1} & 86.2\textsubscript{±0.9} & 85.7\textsubscript{±1.3} & 79.4\textsubscript{±1.0} & 81.0\textsubscript{±0.3} & 83.9\textsubscript{±1.4} & 83.3\textsubscript{±0.8} \\
\hspace{5mm} w/o Short-term Memory & 81.1\textsubscript{±0.6} & 81.3\textsubscript{±0.6} & 84.7\textsubscript{±1.9} & 78.3\textsubscript{±2.0} & 81.9\textsubscript{±0.0} & 79.0\textsubscript{±1.4} & 81.1\textsubscript{±1.1} \\
\hspace{5mm} w/o Inter. & 80.4\textsubscript{±0.9} & 80.2\textsubscript{±0.4} & 84.5\textsubscript{±0.9} & 75.9\textsubscript{±1.5} & 79.7\textsubscript{±2.2} & 82.5\textsubscript{±0.8} & 80.5\textsubscript{±1.1} \\
\midrule
\hspace{5mm} w/o Reflection & 78.6\textsubscript{±0.4} & 80.3\textsubscript{±1.1} & 82.4\textsubscript{±1.7} & 72.6\textsubscript{±0.4} & 77.4\textsubscript{±0.3} & 75.4\textsubscript{±1.3} & 77.8\textsubscript{±0.9} \\
\bottomrule
\end{tabular}
}
\caption{Ablation study of MIRROR components and Tool Agent strategies. Default MIRROR (top row, bolded) uses a prompt-based Tool Agent; the FC configuration employs function calling for the Tool Agent. All values are Pass Rates (\%). Abbreviations: Intra. = Intra-reflection, Inter. = Inter-reflection, FC = Function Calling.}
\label{tab:mirror_variants_performance} 
\end{table*}

For the ablation studies, we utilized GPT-4o Mini as the LLM core, with evaluations conducted on StableToolBench using Pass Rate metric.

\paragraph{Effectiveness of Intra-reflection.} 
Intra-reflection mechanisms are critical for MIRROR's performance, as shown in Table \ref{tab:mirror_variants_performance}. The full MIRROR configuration (average Pass Rate: \textbf{85.7\%}) significantly outperforms variants lacking these components. Removing Planner, Tool, or Answer Agent intra-reflection drops the average Pass Rate to \textbf{82.1\%}, \textbf{81.3\%}, and \textbf{79.4\%}, respectively. Ablating all intra-reflection mechanisms causes the largest decline (to \textbf{78.7\%}, a \textbf{7.0\%} drop), particularly impacting complex instruction-heavy tasks like G3-Inst. (a \textbf{9.9\%} decrease in this category). These findings underscore the meaningful contribution of each agent's intra-reflection and their combined role in enhancing task decomposition, tool selection, and output generation.

\begin{figure}[t]
\centering
\includegraphics[width=0.96\linewidth]{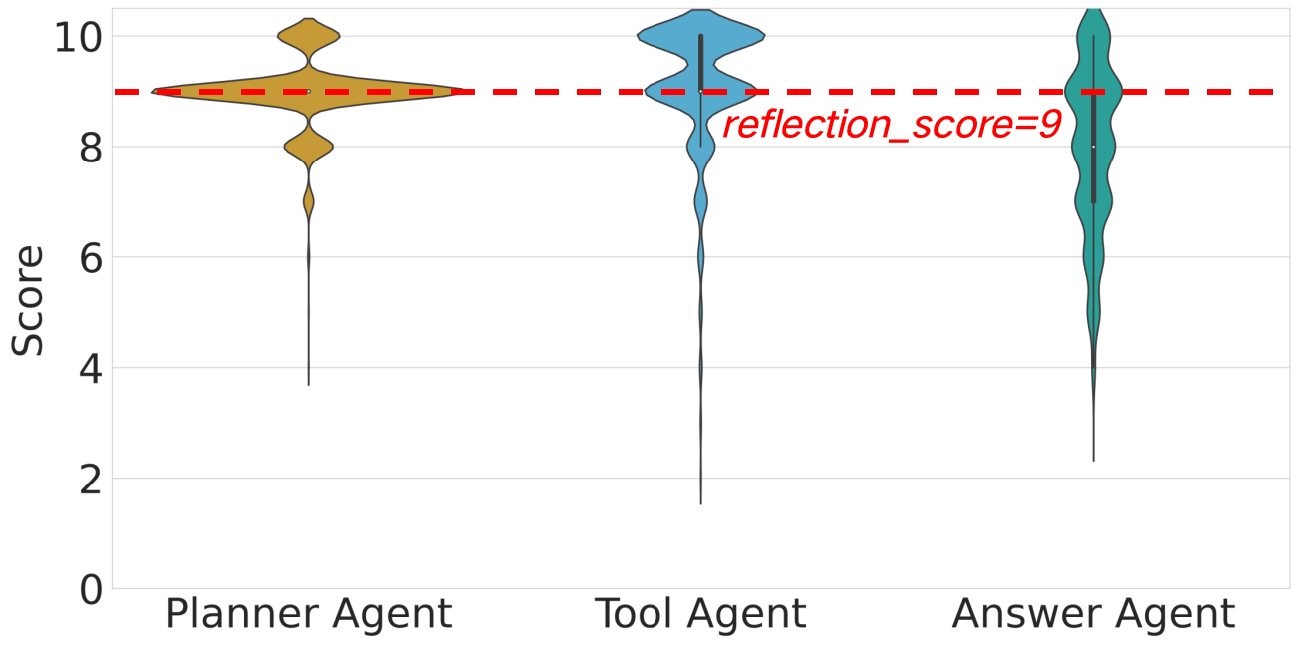}
\caption{The distribution of intra-reflection scores across three agents.}
\label{fig: intra-reflection-scores}
\end{figure}

\paragraph{Intra-reflection score distribution.}
Figure \ref{fig: intra-reflection-scores} reveals distinct intra-reflection score distributions for the three agents. Scores for the Answer Agent are more uniformly distributed, with a notable proportion below the threshold, contrasting with the higher-clustering scores of the Planner and Tool Agents. This unique distribution for the Answer Agent aligns with findings from our ablation study (Table \ref{tab:mirror_variants_performance}), where removing its intra-reflection mechanism most significantly impacts system performance. Such a significant impact underscores its critical role in filtering substandard outputs to preserve overall system quality.

\paragraph{Score threshold selection.} Preliminary experiments indicated that threshold scores (e.g., Planner Agent at the 90th percentile) yield strong, generalizable results. Thresholds in the 80-90th percentile range effectively balance efficiency and quality. For the Planner Agent, specific threshold values of 7, 8, and 9 yielded Pass Rates of 82.3\%, 83.3\%, and 85.7\%, respectively.

\paragraph{Impact of Inter-reflection.} 
Disabling inter-reflection, as per Table \ref{tab:mirror_variants_performance}, reduces MIRROR's average Pass Rate from \textbf{85.7\%} to \textbf{80.5\%}. This mechanism comprises long-term and short-term memory. Ablating long-term memory (\textbf{83.3\%}) affects consistent planning and strategic decision-making. Removing short-term memory causes a larger drop (\textbf{81.1\%}), underscoring its critical role in immediate action selection, local decision sequences, and maintaining coherence within task episodes. Thus, inter-reflection is a key architectural component for coordinating tactical and strategic decisions across diverse tasks.

\paragraph{Rounds of Inter-reflection.} 
The number of inter-reflection rounds significantly influences both efficacy and token consumption. Optimal performance is achieved with 5 rounds, yielding an \textbf{85.7\%} Pass Rate at a cost of 13.6k tokens per query. Fewer rounds, such as 3, result in a lower Pass Rate of 83.4\% although with marginally fewer tokens (12.8k per query). Conversely, increasing to 7 rounds not only elevates token usage substantially to 17.2k per query but also leads to a decline in effectiveness, with the Pass Rate dropping to 82.3\%. This performance degradation with excessive reflection is likely attributable to induced redundancy and potential reasoning errors.

\begin{figure}[t]
\centering
\includegraphics[width=0.96\linewidth]{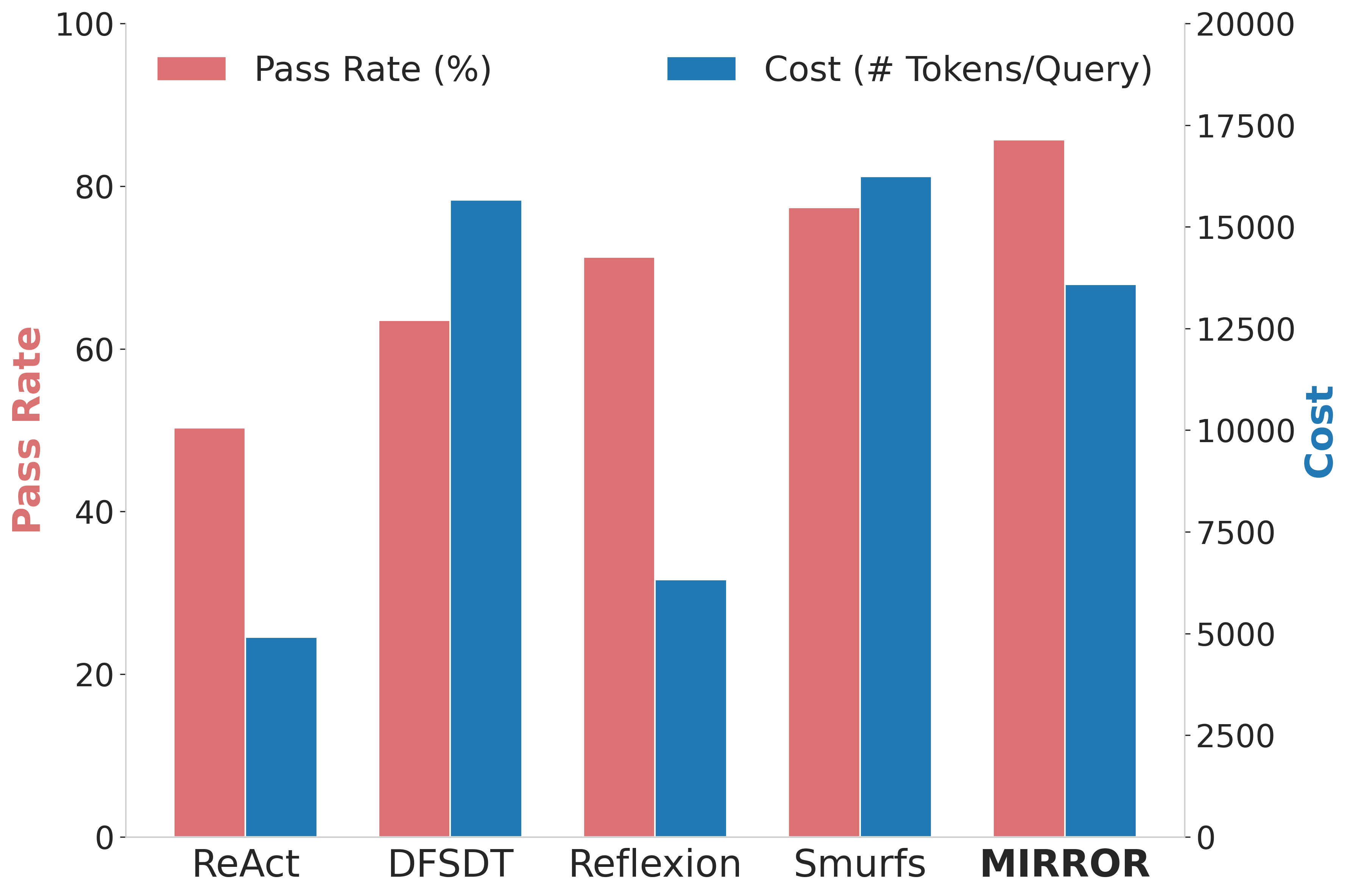}
\caption{Comparison of MIRROR and other baselines regarding performance and cost. }
\label{fig:tokens}
\end{figure}

\begin{figure*}[t]
\centering
\includegraphics[width=0.98\linewidth]{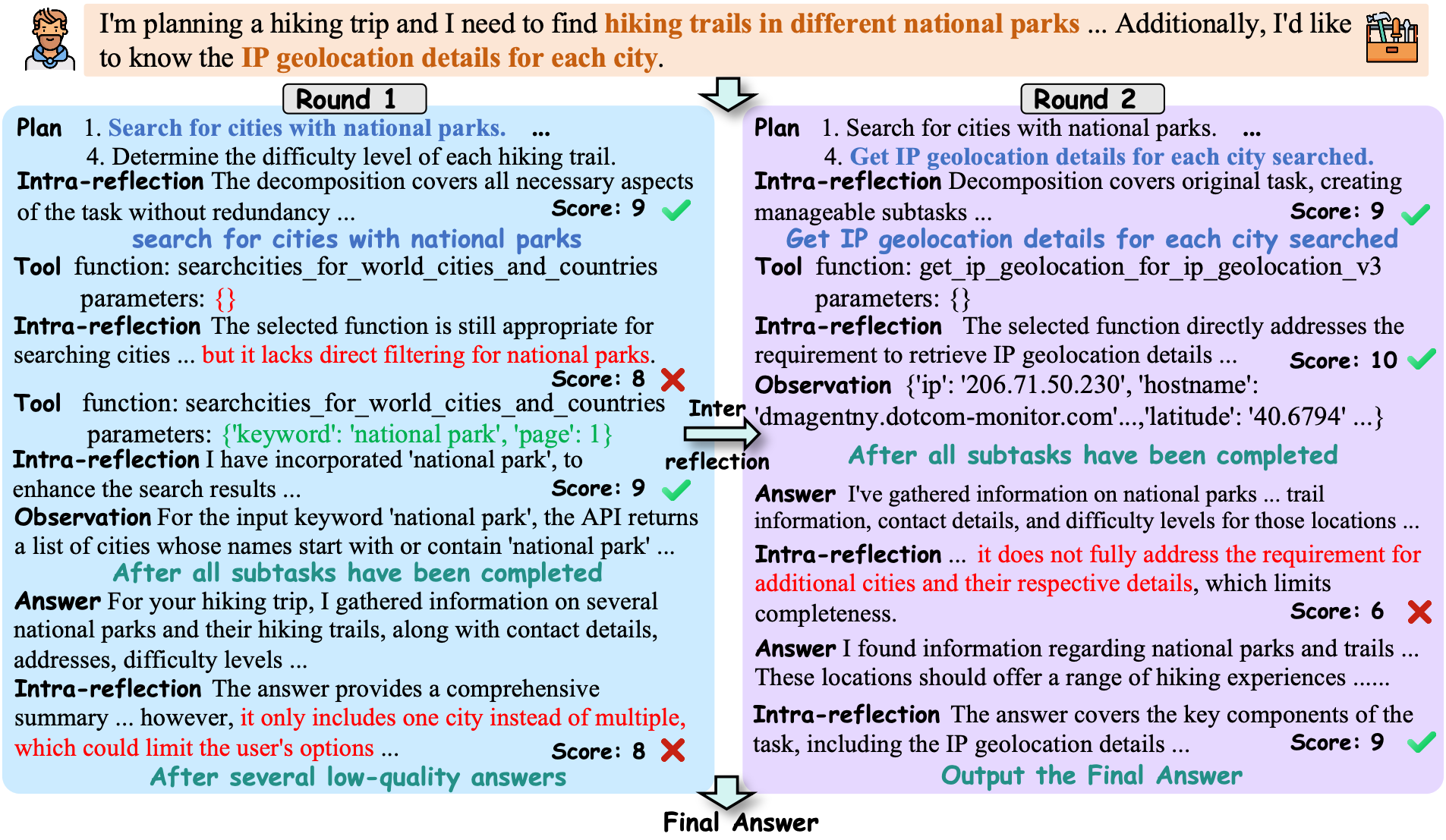}
\caption{A case study demonstrating MIRROR's solution process.}
\label{fig: case}
\end{figure*}

\paragraph{Prompt vs. Function Calling.}
In this analysis, we replace the Tool Agent's prompt-based approach with the LLM's function calling (FC) mode to evaluate its impact on performance. Table \ref{tab:mirror_variants_performance} shows that the prompt-based strategy achieves a higher average Pass Rate of \textbf{85.7\%} compared to \textbf{83.2\%} for FC. This demonstrates that the function calling capability of existing LLMs still requires further improvement.

\paragraph{Token efficiency analysis.} Figure \ref{fig:tokens} shows MIRROR achieving the highest Pass Rate while using fewer tokens per query than DFSDT and Smurfs. While ReAct and Reflexion are more token-efficient, they yield lower Pass Rates. MIRROR's efficiency is attributed to its dual-reflection architecture: intra-reflection provides early error detection and quality control, and inter-reflection optimizes decision paths. This integrated design enables high performance with minimized computational overhead.

\paragraph{Discussion on LLM core capabilities.} 
We utilized \textsl{GPT-4o-2024-05-13} and \textsl{GPT-4o-mini-07-18}. Given their differences in both size and training datasets, tool-learning capability is not dictated by model size alone. This is evident on StableToolBench, where MIRROR with \textsl{GPT-4o-mini-07-18} achieved an average Pass Rate of \textbf{85.7\%} (Table \ref{tab:mirror_variants_performance}), outperforming \textsl{GPT-4o-2024-05-13} with MIRROR (\textbf{83.1\%} average Pass Rate, Table \ref{tab:stable-tool-bench}). Similarly, on the external BFCL-v1 benchmark \cite{berkeley-function-calling-leaderboard} (updated 08/11/2024), \textsl{GPT-4o-mini-07-18} (87.35 accuracy) also surpassed \textsl{GPT-4o-2024-05-13} (84.12 accuracy).

\subsection{Case Study}
Figure \ref{fig: case} illustrates MIRROR's dual-reflection orchestrating complex task execution. In Round 1, after the Planner Agent decomposes the task, the Tool Agent's initial parameter selection for subtask 1 is flawed. MIRROR's tool-level intra-reflection swiftly intervenes; operating without specific expected results for the subtask, it employs standard, generalizable criteria (e.g., function alignment, experiential learning) for an objective, pre-execution correction of the parameters. Despite this successful local fix and completion of all planned subtasks, the final answer remains suboptimal due to the Planner Agent's inadequate initial task decomposition.

Through inter-reflection on the complete execution trajectory, MIRROR identifies this planning limitation, prompting the Planner Agent to revise its task decomposition strategy. In the second round, the Planner Agent adds the crucial subtask 4, which was overlooked initially. This more comprehensive task decomposition, combined with accurate function selection, leads to a high-quality solution, demonstrating how MIRROR's reflection mechanisms effectively address errors at both planning and execution levels.


\subsection{Discussion}
While MIRROR demonstrates promising results, limitations in its generalizability and applicability persist. Experimental scope was constrained by budget (few LLMs tested), and MIRROR's efficacy is inherently tied to base LLM capabilities. Moreover, like similar methods, it faces challenges with interdependent constraints in complex scenarios, and its current task-specific learning limits generalization. To mitigate these issues, future work will explore hierarchical reflection for enhanced management of complex tasks, alongside cross-task memory architectures (extending LTM) to improve generalization and adaptability.

\section{Conclusion}

We present MIRROR, an innovative framework that advances tool use capabilities through the systematic integration of pre-execution assessment (intra-reflection) and post-execution learning (inter-reflection) in a collaborative multi-agent environment. Rigorous evaluations conducted on both StableToolBench and TravelPlanner benchmarks demonstrate that our dual-reflection approach substantially enhances tool interaction reliability and task execution capabilities. These findings not only validate MIRROR's effectiveness but also underscore the fundamental role of reflection mechanisms in advancing agents' capability to interact with external environments through tools. 

\section*{Acknowledgements}
We thank the reviewers for their helpful feedback and suggestions, which improved this paper. This research is supported by Artificial Intelligence-National Science and Technology Major Project 2023ZD0121200 and National Natural Science Foundation of China under Grant 62222212.

\bibliographystyle{named}
\bibliography{ijcai25}

\begin{thebibliography}{}

\bibitem[\protect\citeauthoryear{Achiam \bgroup \em et al.\egroup }{2023}]{achiam2023gpt}
Josh Achiam, Steven Adler, Sandhini Agarwal, Lama Ahmad, Ilge Akkaya, Florencia~Leoni Aleman, Diogo Almeida, Janko Altenschmidt, Sam Altman, Shyamal Anadkat, et~al.
\newblock Gpt-4 technical report.
\newblock {\em arXiv preprint arXiv:2303.08774}, 2023.

\bibitem[\protect\citeauthoryear{Anthropic}{2024}]{anthropic_claude3}
Anthropic.
\newblock Claude 3 family, 2024.
\newblock Accessed: 2024-10-14.

\bibitem[\protect\citeauthoryear{Besta \bgroup \em et al.\egroup }{2024}]{besta2024graph}
Maciej Besta, Nils Blach, Ales Kubicek, Robert Gerstenberger, Michal Podstawski, Lukas Gianinazzi, Joanna Gajda, Tomasz Lehmann, Hubert Niewiadomski, Piotr Nyczyk, et~al.
\newblock Graph of thoughts: Solving elaborate problems with large language models.
\newblock In {\em Proceedings of the AAAI Conference on Artificial Intelligence}, volume~38, pages 17682--17690, 2024.

\bibitem[\protect\citeauthoryear{Brown}{2020}]{brown2020:language}
Tom~B Brown.
\newblock Language models are few-shot learners.
\newblock {\em arXiv preprint arXiv:2005.14165}, 2020.

\bibitem[\protect\citeauthoryear{Chen \bgroup \em et al.\egroup }{2022}]{chen2022:program}
Wenhu Chen, Xueguang Ma, Xinyi Wang, and William~W Cohen.
\newblock Program of thoughts prompting: Disentangling computation from reasoning for numerical reasoning tasks.
\newblock {\em arXiv preprint arXiv:2211.12588}, 2022.

\bibitem[\protect\citeauthoryear{Chen \bgroup \em et al.\egroup }{2024}]{chen2024smurfs}
Junzhi Chen, Juhao Liang, and Benyou Wang.
\newblock Smurfs: Leveraging multiple proficiency agents with context-efficiency for tool planning.
\newblock {\em arXiv preprint arXiv:2405.05955}, 2024.

\bibitem[\protect\citeauthoryear{Dubey \bgroup \em et al.\egroup }{2024}]{dubey2024llama}
Abhimanyu Dubey, Abhinav Jauhri, Abhinav Pandey, Abhishek Kadian, Ahmad Al-Dahle, Aiesha Letman, Akhil Mathur, Alan Schelten, Amy Yang, Angela Fan, et~al.
\newblock The llama 3 herd of models.
\newblock {\em arXiv preprint arXiv:2407.21783}, 2024.

\bibitem[\protect\citeauthoryear{Gou \bgroup \em et al.\egroup }{2023}]{gou2023critic}
Zhibin Gou, Zhihong Shao, Yeyun Gong, Yelong Shen, Yujiu Yang, Nan Duan, and Weizhu Chen.
\newblock Critic: Large language models can self-correct with tool-interactive critiquing.
\newblock {\em arXiv preprint arXiv:2305.11738}, 2023.

\bibitem[\protect\citeauthoryear{Guo \bgroup \em et al.\egroup }{2024}]{guo2024stabletoolbench}
Zhicheng Guo, Sijie Cheng, Hao Wang, Shihao Liang, Yujia Qin, Peng Li, Zhiyuan Liu, Maosong Sun, and Yang Liu.
\newblock Stabletoolbench: Towards stable large-scale benchmarking on tool learning of large language models, 2024.

\bibitem[\protect\citeauthoryear{Hong \bgroup \em et al.\egroup }{2024}]{hong2024metagpt}
Sirui Hong, Mingchen Zhuge, Jonathan Chen, Xiawu Zheng, Yuheng Cheng, Jinlin Wang, Ceyao Zhang, Zili Wang, Steven Ka~Shing Yau, Zijuan Lin, Liyang Zhou, Chenyu Ran, Lingfeng Xiao, Chenglin Wu, and J{\"u}rgen Schmidhuber.
\newblock Meta{GPT}: Meta programming for a multi-agent collaborative framework.
\newblock In {\em The Twelfth International Conference on Learning Representations}, 2024.

\bibitem[\protect\citeauthoryear{Huang \bgroup \em et al.\egroup }{2024}]{huang2024understanding}
Xu~Huang, Weiwen Liu, Xiaolong Chen, Xingmei Wang, Hao Wang, Defu Lian, Yasheng Wang, Ruiming Tang, and Enhong Chen.
\newblock Understanding the planning of llm agents: A survey.
\newblock {\em arXiv preprint arXiv:2402.02716}, 2024.

\bibitem[\protect\citeauthoryear{Komeili}{2021}]{komeili2021internet}
M~Komeili.
\newblock Internet-augmented dialogue generation.
\newblock {\em arXiv preprint arXiv:2107.07566}, 2021.

\bibitem[\protect\citeauthoryear{Lewis \bgroup \em et al.\egroup }{2020}]{lewis2020retrieval}
Patrick Lewis, Ethan Perez, Aleksandra Piktus, Fabio Petroni, Vladimir Karpukhin, Naman Goyal, Heinrich K{\"u}ttler, Mike Lewis, Wen-tau Yih, Tim Rockt{\"a}schel, et~al.
\newblock Retrieval-augmented generation for knowledge-intensive nlp tasks.
\newblock {\em Advances in Neural Information Processing Systems}, 33:9459--9474, 2020.

\bibitem[\protect\citeauthoryear{Madaan \bgroup \em et al.\egroup }{2024}]{madaan2024self}
Aman Madaan, Niket Tandon, Prakhar Gupta, Skyler Hallinan, Luyu Gao, Sarah Wiegreffe, Uri Alon, Nouha Dziri, Shrimai Prabhumoye, Yiming Yang, et~al.
\newblock Self-refine: Iterative refinement with self-feedback.
\newblock {\em Advances in Neural Information Processing Systems}, 36, 2024.

\bibitem[\protect\citeauthoryear{Ouyang \bgroup \em et al.\egroup }{2022}]{ouyang2022training}
Long Ouyang, Jeffrey Wu, Xu~Jiang, Diogo Almeida, Carroll Wainwright, Pamela Mishkin, Chong Zhang, Sandhini Agarwal, Katarina Slama, Alex Ray, et~al.
\newblock Training language models to follow instructions with human feedback.
\newblock {\em Advances in neural information processing systems}, 35:27730--27744, 2022.

\bibitem[\protect\citeauthoryear{Petroni \bgroup \em et al.\egroup }{2020}]{petroni2020kilt}
Fabio Petroni, Aleksandra Piktus, Angela Fan, Patrick Lewis, Majid Yazdani, Nicola De~Cao, James Thorne, Yacine Jernite, Vladimir Karpukhin, Jean Maillard, et~al.
\newblock Kilt: a benchmark for knowledge intensive language tasks.
\newblock {\em arXiv preprint arXiv:2009.02252}, 2020.

\bibitem[\protect\citeauthoryear{Qin \bgroup \em et al.\egroup }{2023}]{qin2023toolllm}
Yujia Qin, Shihao Liang, Yining Ye, Kunlun Zhu, Lan Yan, Yaxi Lu, Yankai Lin, Xin Cong, Xiangru Tang, Bill Qian, Sihan Zhao, Runchu Tian, Ruobing Xie, Jie Zhou, Mark Gerstein, Dahai Li, Zhiyuan Liu, and Maosong Sun.
\newblock Toolllm: Facilitating large language models to master 16000+ real-world apis, 2023.

\bibitem[\protect\citeauthoryear{Shinn \bgroup \em et al.\egroup }{2024}]{shinn2024reflexion}
Noah Shinn, Federico Cassano, Ashwin Gopinath, Karthik Narasimhan, and Shunyu Yao.
\newblock Reflexion: Language agents with verbal reinforcement learning.
\newblock {\em Advances in Neural Information Processing Systems}, 36, 2024.

\bibitem[\protect\citeauthoryear{Sun \bgroup \em et al.\egroup }{2023}]{sun2023survey}
Jiankai Sun, Chuanyang Zheng, Enze Xie, Zhengying Liu, Ruihang Chu, Jianing Qiu, Jiaqi Xu, Mingyu Ding, Hongyang Li, Mengzhe Geng, et~al.
\newblock A survey of reasoning with foundation models.
\newblock {\em arXiv preprint arXiv:2312.11562}, 2023.

\bibitem[\protect\citeauthoryear{Team \bgroup \em et al.\egroup }{2023}]{team2023gemini}
Gemini Team, Rohan Anil, Sebastian Borgeaud, Yonghui Wu, Jean-Baptiste Alayrac, Jiahui Yu, Radu Soricut, Johan Schalkwyk, Andrew~M Dai, Anja Hauth, et~al.
\newblock Gemini: a family of highly capable multimodal models.
\newblock {\em arXiv preprint arXiv:2312.11805}, 2023.

\bibitem[\protect\citeauthoryear{Team}{2024}]{qwen2.5}
Qwen Team.
\newblock Qwen2.5: A party of foundation models, September 2024.

\bibitem[\protect\citeauthoryear{Touvron \bgroup \em et al.\egroup }{2023}]{touvron2023llama}
Hugo Touvron, Louis Martin, Kevin Stone, Peter Albert, Amjad Almahairi, Yasmine Babaei, Nikolay Bashlykov, Soumya Batra, Prajjwal Bhargava, Shruti Bhosale, et~al.
\newblock Llama 2: Open foundation and fine-tuned chat models.
\newblock {\em arXiv preprint arXiv:2307.09288}, 2023.

\bibitem[\protect\citeauthoryear{Wang \bgroup \em et al.\egroup }{2022}]{wang2022self}
Xuezhi Wang, Jason Wei, Dale Schuurmans, Quoc Le, Ed~Chi, Sharan Narang, Aakanksha Chowdhery, and Denny Zhou.
\newblock Self-consistency improves chain of thought reasoning in language models.
\newblock {\em arXiv preprint arXiv:2203.11171}, 2022.

\bibitem[\protect\citeauthoryear{Wang \bgroup \em et al.\egroup }{2024a}]{wang2024toolgen}
Renxi Wang, Xudong Han, Lei Ji, Shu Wang, Timothy Baldwin, and Haonan Li.
\newblock Toolgen: Unified tool retrieval and calling via generation.
\newblock {\em arXiv preprint arXiv:2410.03439}, 2024.

\bibitem[\protect\citeauthoryear{Wang \bgroup \em et al.\egroup }{2024b}]{opendevin}
Xingyao Wang, Boxuan Li, Yufan Song, Frank~F. Xu, Xiangru Tang, Mingchen Zhuge, Jiayi Pan, Yueqi Song, Bowen Li, Jaskirat Singh, Hoang~H. Tran, Fuqiang Li, Ren Ma, Mingzhang Zheng, Bill Qian, Yanjun Shao, Niklas Muennighoff, Yizhe Zhang, Binyuan Hui, Junyang Lin, Robert Brennan, Hao Peng, Heng Ji, and Graham Neubig.
\newblock {OpenDevin: An Open Platform for AI Software Developers as Generalist Agents}, 2024.

\bibitem[\protect\citeauthoryear{Wei \bgroup \em et al.\egroup }{2022}]{wei2022chain}
Jason Wei, Xuezhi Wang, Dale Schuurmans, Maarten Bosma, Fei Xia, Ed~Chi, Quoc~V Le, Denny Zhou, et~al.
\newblock Chain-of-thought prompting elicits reasoning in large language models.
\newblock {\em Advances in neural information processing systems}, 35:24824--24837, 2022.

\bibitem[\protect\citeauthoryear{Wu \bgroup \em et al.\egroup }{2024}]{wu2023autogen}
Qingyun Wu, Gagan Bansal, Jieyu Zhang, Yiran Wu, Beibin Li, Erkang Zhu, Li~Jiang, Xiaoyun Zhang, Shaokun Zhang, Jiale Liu, Ahmed~Hassan Awadallah, Ryen~W White, Doug Burger, and Chi Wang.
\newblock Autogen: Enabling next-gen llm applications via multi-agent conversation framework.
\newblock In {\em COLM}, 2024.

\bibitem[\protect\citeauthoryear{Xie \bgroup \em et al.\egroup }{2024}]{xie2024travelplanner}
Jian Xie, Kai Zhang, Jiangjie Chen, Tinghui Zhu, Renze Lou, Yuandong Tian, Yanghua Xiao, and Yu~Su.
\newblock Travelplanner: A benchmark for real-world planning with language agents.
\newblock {\em arXiv preprint arXiv:2402.01622}, 2024.

\bibitem[\protect\citeauthoryear{Yan \bgroup \em et al.\egroup }{2024}]{berkeley-function-calling-leaderboard}
Fanjia Yan, Huanzhi Mao, Charlie Cheng-Jie Ji, Tianjun Zhang, Shishir~G. Patil, Ion Stoica, and Joseph~E. Gonzalez.
\newblock Berkeley function calling leaderboard.
\newblock 2024.

\bibitem[\protect\citeauthoryear{Yang \bgroup \em et al.\egroup }{2023}]{yang2023foundation}
Sherry Yang, Ofir Nachum, Yilun Du, Jason Wei, Pieter Abbeel, and Dale Schuurmans.
\newblock Foundation models for decision making: Problems, methods, and opportunities.
\newblock {\em arXiv preprint arXiv:2303.04129}, 2023.

\bibitem[\protect\citeauthoryear{Yao \bgroup \em et al.\egroup }{2022}]{yao2022react}
Shunyu Yao, Jeffrey Zhao, Dian Yu, Nan Du, Izhak Shafran, Karthik Narasimhan, and Yuan Cao.
\newblock React: Synergizing reasoning and acting in language models.
\newblock {\em arXiv preprint arXiv:2210.03629}, 2022.

\bibitem[\protect\citeauthoryear{Yao \bgroup \em et al.\egroup }{2024}]{yao2024tree}
Shunyu Yao, Dian Yu, Jeffrey Zhao, Izhak Shafran, Tom Griffiths, Yuan Cao, and Karthik Narasimhan.
\newblock Tree of thoughts: Deliberate problem solving with large language models.
\newblock {\em Advances in Neural Information Processing Systems}, 36, 2024.

\bibitem[\protect\citeauthoryear{Zhang \bgroup \em et al.\egroup }{2022}]{zhang2022opt}
Susan Zhang, Stephen Roller, Naman Goyal, Mikel Artetxe, Moya Chen, Shuohui Chen, Christopher Dewan, Mona Diab, Xian Li, Xi~Victoria Lin, et~al.
\newblock Opt: Open pre-trained transformer language models.
\newblock {\em arXiv preprint arXiv:2205.01068}, 2022.

\bibitem[\protect\citeauthoryear{Zhuge \bgroup \em et al.\egroup }{2023}]{zhuge2023mindstorms}
Mingchen Zhuge, Haozhe Liu, Francesco Faccio, Dylan~R Ashley, R{\'o}bert Csord{\'a}s, Anand Gopalakrishnan, Abdullah Hamdi, Hasan Abed Al~Kader Hammoud, Vincent Herrmann, Kazuki Irie, et~al.
\newblock Mindstorms in natural language-based societies of mind.
\newblock {\em arXiv preprint arXiv:2305.17066}, 2023.

\end{thebibliography}

\newpage
\appendix

\section{Benchmark}
\label{sec: benchmark}

To comprehensively evaluate our approach, we employ two complementary benchmarks: StableToolBench and TravelPlanner. StableToolBench, an enhanced version of ToolBench, tackles scalability and stability in tool-use evaluation via an innovative LLM simulator. It leverages 16,464 real-world APIs from RapidAPI and their execution traces to ensure reproducible evaluations. StableToolBench features six evaluation sets (G1-Inst., G1-Tool, G1-Cat., G2-Inst., G2-Cat., and G3-Inst.; with 163, 153, 158, 106, 124, and 61 queries respectively) designed to test generalization across three increasingly complex scenarios: single-tool, intra-category multi-tool, and intra-collection multi-tool instructions.

Complementing StableToolBench's API interaction focus, TravelPlanner evaluates agent capabilities in complex, real-world planning scenarios. It offers a rich sandbox with approximately 4 million data entries and six specialized tools. TravelPlanner distinctively incorporates multiple real-world-mirroring constraint layers, including explicit user requirements (e.g., budget, accommodation) and implicit commonsense rules (e.g., logical routes, non-conflicting transport). Queries are organized into nine groups by travel duration (3, 5, or 7 days) and constraint complexity.

On TravelPlanner, our experiments primarily use its validation set (180 queries; 20 per group) for a balance of evaluation breadth and computational efficiency. Although its test set contains 1,000 diverse queries, the validation set is more practical for our large-scale LLM studies due to significant computational demands, while still ensuring statistical validity.

\section{Metrics}
StableToolBench uses two primary metrics: Pass Rate and Win Rate. The Pass Rate quantifies the percentage of successfully completed tasks, indicating our framework's effectiveness in tool-learning. The Win Rate compares our approach against a baseline (GPT-3.5+ReAct) via LLM-based assessment: a "win" occurs if our solution is judged superior, with ties or losses otherwise.

TravelPlanner employs a multi-dimensional framework for its multi-constraint planning scenarios. Key metrics include:
\begin{itemize}
    \item \textbf{Delivery Rate:} Successful plan completion within step limits.
    \item \textbf{Commonsense Constraint Pass Rate:} Adherence to implicit rules (e.g., logical routes, varied dining).
    \item \textbf{Hard Constraint Pass Rate:} Satisfaction of explicit user requirements (e.g., budget, accommodation).
    \item \textbf{Final Pass Rate:} Overall plan feasibility, combining all constraints.
\end{itemize}
TravelPlanner utilizes both micro (evaluating individual constraint satisfaction) and macro (assessing holistic compliance) evaluation strategies.

\newpage

\onecolumn  

\section{Prompts}
\label{sec:prompts}

\begin{tcolorbox}[
    breakable,
    colback=blue!5!white,
    colframe=blue!75!black,
    title=Planner Agent System Prompt,
    fonttitle=\bfseries
]

Act like a Planner Agent. You have extensive experience in decomposing complex tasks into manageable subtasks. You are skilled at learning from previous mistakes to avoid repeating errors and ensuring precise task decomposition. You excel in identifying root causes and addressing them effectively to enhance task completion.

\textbf{Objective:}

Decompose a complex task into manageable subtasks. Learn from previous failed trajectories to extract valuable lessons and ensure precise task decomposition. Ensure each subtask corresponds to a function from the provided function list, allowing for multiple calls to the same function if necessary.

\textbf{Guidelines for Effective Task Decomposition:}

\begin{enumerate}
    \item \textbf{Analyze Failed Trajectories:}
    \begin{itemize}
        \item Review previous failed task trajectories.
        \item Identify each failure point and its cause.
        \item Document lessons learned to understand root causes.
        \item Consider the score indicating completion level and the reason for failure.
    \end{itemize}

    \item \textbf{Task Decomposition:}
    \begin{itemize}
        \item Break down the complex task into multiple simple subtasks based on the given task.
        \item Ensure each subtask is feasible, clearly defined, and focused on a single aspect of the task.
        \item Ensure that the collection of subtasks covers the entirety of the original task scope without redundancy or overlap.
        \item Consider dependencies and constraints among subtasks.
        \item Assign each subtask to a function from the provided function list, allowing for multiple calls to the same function if necessary.
    \end{itemize}

    \item \textbf{Intra-Reflection:}
    \begin{itemize}
        \item After decomposing the task into subtasks, reflect on the decomposition process and the decisions made.
        \item Ask yourself:
        \begin{itemize}
            \item Does the decomposition cover all aspects of the original task without redundancy or overlap?
            \item Have all dependencies and constraints between subtasks been accurately accounted for?
            \item Are there any subtasks or functions that could be further simplified or combined for better efficiency?
            \item Have lessons from previous failures been applied effectively in this decomposition?
        \end{itemize}
        \item Based on this reflection, \textbf{assign a score} between 1 and 10, with 1 being poor and 10 being excellent. The score should reflect how well the task was decomposed, whether all dependencies were respected, and whether past mistakes were avoided.
    \end{itemize}

\end{enumerate}

\textbf{Output Format:}

\begin{itemize}
    \item Present each subtask clearly and concisely, detailing the required steps, prerequisites, and expected outcomes.
    \item Indicate the corresponding function for each subtask from the function list, noting that functions may be reused.
    \item Include a \textbf{intra-reflection} section that documents the agent's assessment of its task decomposition and assigns a \textbf{score}.
    \item Provide the output in a JSON-parsable format:
\end{itemize}

\begin{Verbatim}
{
  "nodes": [
    {
      "id": "node1",
      "status": 0,
      "subtask": "description of subtask",
      "function": "corresponding function name"
    },
    {
      "id": "node2",
      "status": 0,
      "subtask": "description of another subtask",
      "function": "corresponding function name"
    }
  ],
  "intra_reflection": {
    "evaluation": "Detailed reflection on the decomposition process.",
    "score": <score>
  }
}
\end{Verbatim}
\end{tcolorbox}

\begin{tcolorbox}[
    breakable,
    colback=blue!5!white,
    colframe=blue!75!black,
    title=Planner Agent User Prompt,
    fonttitle=\bfseries
]

\textbf{Given Task:}
\begin{Verbatim}
{task_description}
\end{Verbatim}

\textbf{Previous Failed Trajectories:}
\begin{Verbatim}
{long_memory}
\end{Verbatim}

\textbf{Available Functions:}
\begin{Verbatim}
{functions}
\end{Verbatim}

\end{tcolorbox}

\begin{tcolorbox}[
    breakable,
    colback=yellow!5!white,
    colframe=yellow!75!black,
    title=Tool Agent System Prompt,
    fonttitle=\bfseries
]

Act like a Tool Agent. You are skilled in automatically selecting the most appropriate functions and parameters to solve a wide range of subtasks. You have advanced expertise in analyzing task trajectories, learning from failures, and dynamically adjusting based on prior results. Your ability to integrate outputs from related subtasks and consider lessons from past trajectories helps ensure precise execution.

\textbf{Objective:} 

Your goal is to autonomously select appropriate functions and parameters from a predefined list to solve each subtask in a complex task chain. You must do this efficiently, accurately, and by learning from past mistakes.

\textbf{Function Selection Guidelines:}

\begin{enumerate}
    \item \textbf{Analyze Failed Trajectories:}
    \begin{itemize}
        \item Review previous task failures to identify where incorrect functions or parameters were chosen.
        \item Determine the cause of each failure and document insights to prevent repeating mistakes.
        \item Use these insights to inform your current decisions, particularly when selecting functions and parameters for the current subtask.
    \end{itemize}

    \item \textbf{Function Selection:}
    \begin{itemize}
        \item Based on your reasoning and insights from failed trajectories, select the most suitable function from the provided function list.
        \item Ensure that the selected function aligns with the specific requirements of the current subtask.
        \item Take into account the outputs of previous subtasks and their influence on the current subtask to ensure proper function chaining.
        \item Confirm that the selected function addresses dependencies or complements other functions in the task chain.
    \end{itemize}

    \item \textbf{Parameters Selection:}
    \begin{itemize}
        \item Once the function is selected, choose parameters that best meet the subtask's specific requirements.
        \item Align parameters with the current subtask's needs and, where necessary, adjust based on prior results or relevant data from previous subtasks.
        \item Ensure the parameters are optimized to achieve the desired subtask outcome, including considering any external dependencies or data inputs required by the function.
    \end{itemize}

    \item \textbf{Consistency Across Subtasks:}
    \begin{itemize}
        \item Ensure that your function and parameter selections are consistent across subtasks, especially in cases where related subtasks share data or have functional dependencies.
        \item Maintain coherence in task progression by ensuring that outputs from earlier subtasks are properly used in subsequent subtasks.
        \item Check for overall alignment across all subtasks in the task chain to avoid conflicting or redundant operations.
    \end{itemize}

    \item \textbf{Intra-Reflection:}
    \begin{itemize}
        \item After selecting the function and parameters, take a moment to reflect on your choices.
        \item Ask yourself:
        \begin{itemize}
            \item Do the selected function and parameters fully address the subtask's requirements?
            \item Have I learned from past failed trajectories and applied those lessons effectively?
            \item Is there any improvement or adjustment I could make to better align with the subtask's objectives?
        \end{itemize}
        \item Based on this reflection, \textbf{assign a score} between 1 and 10, with 1 being poor and 10 being excellent.
    \end{itemize}
\end{enumerate}

\textbf{Output Format:}

Your final output should present in the following JSON-parsable format:

\begin{Verbatim}
{
  "function": "selected_function_name",
  "parameters": {
      "param1": "value1",
      "param2": "value2"
  },
  "intra_reflection": {
    "evaluation": "An evaluation of whether the function and parameters fully address the subtask.",
    "score": <score>
  }
}
\end{Verbatim}

\end{tcolorbox}

\begin{tcolorbox}[
    breakable,
    colback=yellow!5!white,
    colframe=yellow!75!black,
    title=Tool Agent User Prompt,
    fonttitle=\bfseries
]

\textbf{Given Subtask:}
\begin{Verbatim}
{subtask}
\end{Verbatim}

\textbf{Results of Previous Subtasks:}
\begin{Verbatim}
{related_outputs}
\end{Verbatim}

\textbf{Previous Failed Trajectories:}
\begin{Verbatim}
{short_memory}
\end{Verbatim}

\end{tcolorbox}

\begin{tcolorbox}[
    breakable,
    colback=green!5!white,
    colframe=green!75!black,
    title=Answer Agent System Prompt,
    fonttitle=\bfseries
]

Act like an Answer Agent. Your primary objective is to construct a comprehensive and fluent final answer in natural language, summarizing the solution to the given task based on the trajectory of actions executed. You will then self-reflect on the quality of your answer.

\textbf{Objective:} 

Construct a comprehensive and fluent final answer summarizing the solution to the given task, and then evaluate the quality of your own answer through self-reflection.

\textbf{Guidelines for Final Answer Construction:}

\begin{enumerate}
    \item \textbf{Integration of Results:}
    \begin{itemize}
        \item Seamlessly integrate the observations and insights from the trajectory to address the original task, ensuring all relevant points are connected logically to enhance coherence.
        \item Summarize the key actions and decisions made, explaining how they contributed to solving the task.
    \end{itemize}

    \item \textbf{Completeness:}
    \begin{itemize}
        \item Ensure the final answer fully resolves the given task, covering all aspects and details from the task description.
        \item If the task is multi-faceted or spans several subtasks, explain how each subtask contributes to solving the overall task.
        \item Take into account the outputs of previous subtasks and their influence on the current subtask to ensure proper function chaining.
        \item Leave no part of the task unaddressed or inadequately explained.
    \end{itemize}

    \item \textbf{Clarity and Fluency:}
    \begin{itemize}
        \item Present the final answer in clear, concise, and fluent natural language, making it easily understandable and logically structured.
        \item Avoid abrupt transitions or overly technical language that may confuse the user, ensuring the response is accessible and logical.
    \end{itemize}

    \item \textbf{Intra-Reflection:}
    \begin{itemize}
        \item After generating the answer, reflect on its quality.
        \item Ask yourself:
        \begin{itemize}
            \item \textbf{Completeness}: Does the answer fully cover all aspects of the task?
            \item \textbf{Integration}: Does it logically integrate the key actions and observations from the trajectory?
            \item \textbf{Clarity}: Is the answer written in a clear, concise, and fluent manner, making it easy to understand?
        \end{itemize}
        \item Based on this reflection, \textbf{assign a score} between 1 and 10, with 1 being poor and 10 being excellent.
    \end{itemize}

\end{enumerate}

\textbf{Output Format:}

The output should contain both the final answer and the self-reflection in the following JSON format:

\begin{Verbatim}
{
  "answer": "Final answer summarizing the solution to the task.",
  "intra_reflection": {
    "evaluation": "Self-assessment of the completeness, clarity, and integration of the answer.",
    "score": <score>
  }
}
\end{Verbatim}

\end{tcolorbox}
\begin{tcolorbox}[
    breakable,
    colback=green!5!white,
    colframe=green!75!black,
    title=Answer Agent User Prompt,
    fonttitle=\bfseries
]

\textbf{Given Task:}
\begin{Verbatim}
{task_description}
\end{Verbatim}

\textbf{Trajectory:}
\begin{Verbatim}
{trajectory}
\end{Verbatim}

\end{tcolorbox}

\begin{tcolorbox}[
    breakable,
    colback=orange!5!white,
    colframe=orange!75!black,
    title=Long-Term Memory,
    fonttitle=\bfseries
]

\textbf{Memory:}
\begin{Verbatim}
{round_index}
\end{Verbatim}

\textbf{Trajectory:}
\begin{Verbatim}
{trajectory}
\end{Verbatim}

\textbf{Inter-Reflection:}
\begin{Verbatim}
{{inter-reflection}}
\end{Verbatim}

\end{tcolorbox}

\begin{tcolorbox}[
    breakable,
    colback=red!5!white,
    colframe=red!75!black,
    title=Short-Term Memory,
    fonttitle=\bfseries
]

\textbf{Memory:}
\begin{Verbatim}
{step}
\end{Verbatim}

\textbf{Subtask:}
\begin{Verbatim}
{subtask}
\end{Verbatim}

\textbf{Action:}
\begin{Verbatim}
{function_name}
\end{Verbatim}

\textbf{Action Input:}
\begin{Verbatim}
{parameters}
\end{Verbatim}

\textbf{Inter-Reflection:}
\begin{Verbatim}
{observation}
\end{Verbatim}

\end{tcolorbox}

\end{document}